\documentclass{article}

\usepackage{amsmath}
\usepackage{amssymb}
\usepackage{amsthm}
\usepackage{graphicx}
\newtheorem{proposition}{Proposition}
\newtheorem{lemma}{Lemma}
\PassOptionsToPackage{numbers,compress}{natbib}
\usepackage[preprint]{neurips_2026}
\usepackage{multirow}
\usepackage[utf8]{inputenc} 
\usepackage[T1]{fontenc}    
\usepackage{hyperref}       
\usepackage{url}            
\usepackage{booktabs}       
\usepackage{amsfonts}       
\usepackage{nicefrac}       
\usepackage{microtype}      

\title{RetinexGS: Retinex-Inspired Gaussian Splatting for Low-Light RAW Reconstruction}
\author{
  Yulang Fei \qquad Huaiyuan Weng \qquad Chul Min Yeum \\
  University of Waterloo \\
  Waterloo, Ontario, Canada \\
  \texttt{\{yulang.fei, huaiyuan.weng, cmyeum\}@uwaterloo.ca}
}

\begin{document}

\maketitle

\begin{abstract}
Standard 3D Gaussian Splatting (3DGS) pipelines struggle with severely underexposed images, producing unstable colors, floating artifacts, and incomplete reconstructions in dark regions. To better account for noise in these low-light observations, RAW images offer a natural basis for noise-aware reconstruction by preserving linear sensor measurements and providing access to noise metadata. We present \textbf{RetinexGS}, a RAW-domain 3DGS framework that combines Retinex-inspired appearance modeling with sensor-informed optimization for low-light novel-view synthesis. We factorize each Gaussian’s color into a view-independent base color and a shared, position- and view-dependent modulation field, regularizing the two components separately to constrain appearance learning from noisy observations. We use DNG noise metadata to weight the reconstruction loss for signal-dependent noise, while distortion-aware rendering avoids spatially resampling RAW targets. To address sparse geometry in dark regions, we introduce depth-guided initialization. We evaluate RetinexGS on three public RawNeRF scenes and two custom-captured low-light RAW scenes. With all methods trained on the same RAW inputs for each scene, RetinexGS outperforms LE3D and RawNeRF in mean sRGB PSNR by 2.18 dB and 2.71 dB respectively, across the five scenes, while supporting real-time novel-view rendering.

\end{abstract}
\begin{figure*}[t]
    \centering
    \includegraphics[width=\textwidth]{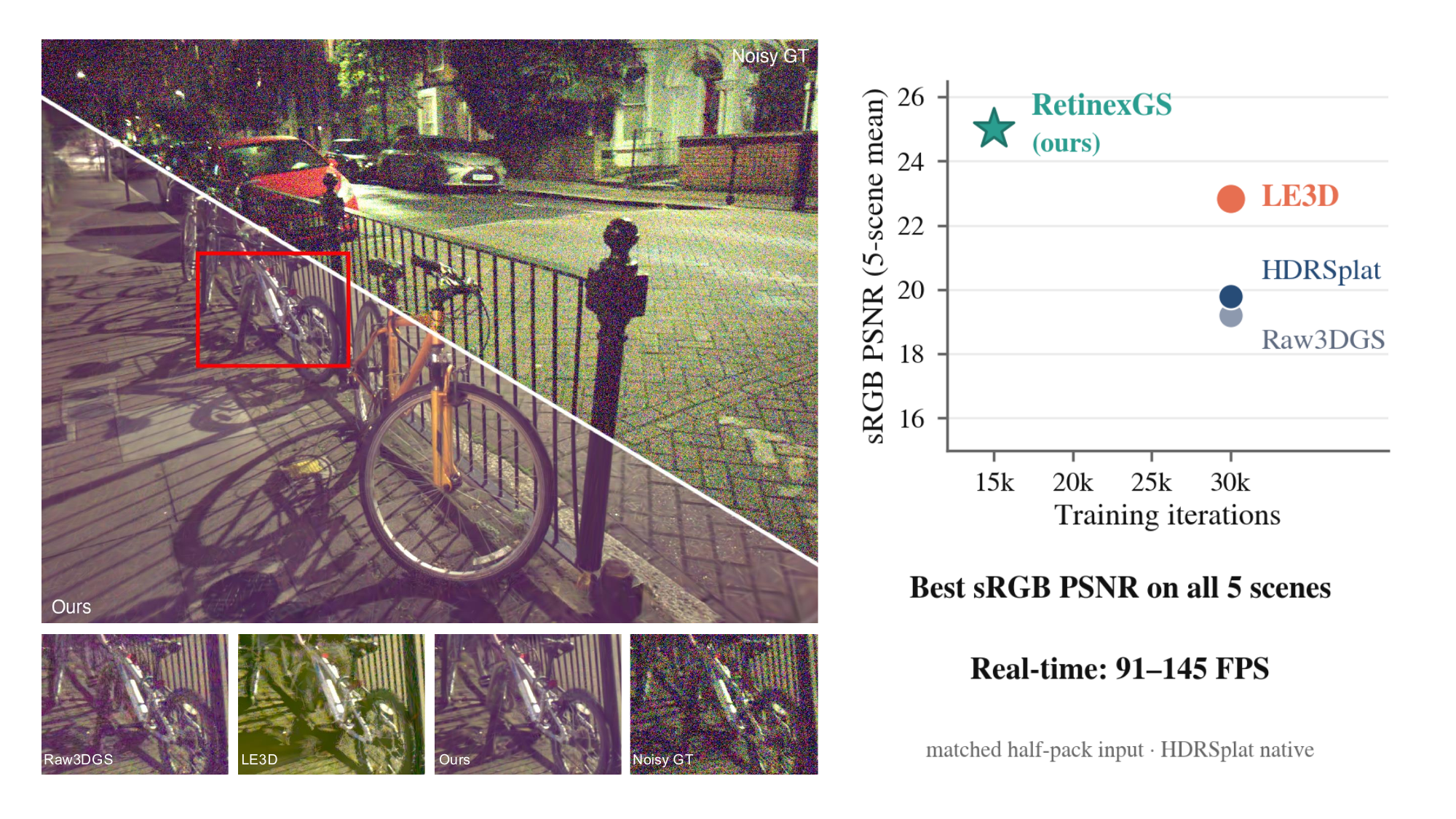}
    \caption{
    \textbf{RetinexGS enables high-quality and efficient low-light RAW novel-view synthesis.}
    \textbf{Left:} On a noisy held-out view, RetinexGS suppresses noise-driven artifacts while preserving thin structures and fine details; the crops compare RawGS, LE3D, RetinexGS, and the observed noisy target.
    \textbf{Right:} RetinexGS achieves the highest affine-sRGB PSNR on each of the five evaluated scenes and the highest five-scene mean after only $15{,}000$ training iterations, compared with $30{,}000$ for the Gaussian baselines. RetinexGS, LE3D, and Raw3DGS receive identical half-pack inputs; HDRSplat retains its native PMRID preprocessing. The five-scene Raw3DGS control is reported in Appendix~\ref{app:raw3dgs}.
    RetinexGS renders at $91$--$145$ FPS at $2016\times1512$ resolution on a single RTX~4090.
    }
    \label{fig:teaser}
\end{figure*}

\section{Introduction}
\label{sec:intro}

Novel-view synthesis (NVS) in low-light environments remains challenging for modern 3D representations. Standard 3D Gaussian Splatting (3DGS)~\cite{kerbl20233dgs} performs well when structure-from-motion (SfM)~\cite{schoenberger2016sfm} provides reliable initialization and well-exposed sRGB images provide stable photometric supervision. Under severely low-light conditions, feature correspondences become sparse or unreliable in dark and texture-poor regions, while sensor noise obscures weak image signals. The resulting sparse point-cloud initialization and noisy supervision can lead to incomplete geometry, unstable colors, and floating artifacts in the reconstructed scene.

A natural alternative is to train 3DGS directly on RAW images, which retain high-bit-depth, approximately linear sensor measurements before image signal processing (ISP) and thus preserve information that may be lost in sRGB images. When RAW images are stored in the Digital Negative (DNG) format~\cite{adobe2012dng}, an open container for RAW sensor data, their metadata may additionally provide per-channel noise parameters (the \texttt{NoiseProfile} tag) that can inform noise-aware optimization. RawNeRF~\cite{mildenhall2022rawnerf} demonstrated that multi-view optimization can reconstruct scenes directly from noisy RAW observations captured in near darkness, and LE3D~\cite{jin2024le3d} extended RAW-domain reconstruction to real-time Gaussian rendering. Building on these advances, we introduce a structured appearance representation to constrain Gaussian color estimation under noisy RAW supervision.

Retinex theory~\cite{land1971lightness} models appearance as the product of reflectance and illumination. Inspired by this principle, we structure Gaussian appearance multiplicatively in linear RAW space: each Gaussian carries a view-independent base color, which is multiplied by a modulation that a compact, scene-shared field predicts from the Gaussian's position and the viewing direction. Because the field is shared across the scene, it captures smooth spatial and view-dependent variation, while per-Gaussian color detail is carried by the base factor. The factorization alone is ambiguous, since scale can be traded freely between the two factors; its value lies in allowing each factor to be constrained separately. Rather than fitting a single per-Gaussian color directly to noisy observations, we can therefore impose priors suited to each factor.

We propose \textbf{RetinexGS}, a RAW-domain framework that implements this idea through factor-specific regularization and sensor-informed optimization. We constrain the modulation scale and encourage local consistency of the base colors (Sec.~\ref{sec:retinex}), while fitting packed RAW observations with channel-wise noise weights derived from the DNG \texttt{NoiseProfile}. We handle lens distortion within the rasterizer, enabling supervision directly on the packed RAW grid without spatially resampling the targets. To improve geometric coverage, we augment SfM initialization with monocular-depth samples in dark regions with insufficient SfM points. Experiments on public and newly captured low-light RAW scenes demonstrate improved reconstruction quality under both native-pipeline and matched-input comparisons, while retaining real-time rendering. Fig.~\ref{fig:teaser} provides a visual overview.

Our contributions are:
Our contributions are:
\begin{itemize}
    \item \textbf{Factored appearance for noisy RAW supervision.} To constrain Gaussian color estimation under noise, we represent each Gaussian's color as the product of a view-independent base color and a modulation predicted by a scene-shared field from the Gaussian's position and the viewing direction. This factorization lets us regularize the base color and the modulation separately. The product is formed per Gaussian before alpha compositing.

    \item \textbf{Sensor-informed RAW optimization.} We exploit the camera's own sensor and lens properties during training. Each RAW channel is weighted by a noise level derived from the DNG \texttt{NoiseProfile}. Lens distortion is modeled inside the rasterizer, so the loss is computed on the original packed RAW pixels without resampling them. In addition, we use monocular depth to add initialization points in dark regions where SfM points are sparse.

    \item \textbf{Evaluation on low-light RAW scenes.} We demonstrate improved reconstruction quality on public and newly captured low-light RAW scenes, supported by ablations of the appearance model and of where the product is formed. Forming the product per Gaussian achieves quality comparable to multiplying rendered images, while roughly halving training time in our implementation.
\end{itemize}

\section{Related Work}
\label{sec:related}

\paragraph{3D Gaussian Splatting and appearance.}
3DGS~\cite{kerbl20233dgs} represents scenes as anisotropic Gaussian primitives rendered through differentiable alpha compositing, enabling high-quality novel-view synthesis at real-time rates. Subsequent work improves geometry~\cite{huang20242dgs,yu2024gof}, anti-aliasing~\cite{yu2024mipsplatting}, and training efficiency~\cite{mallick2024taming3dgs}. Standard 3DGS represents view-dependent color using spherical harmonics coefficients. Extensions incorporate shading functions, material properties, and illumination models for inverse rendering and relighting~\cite{gao2024relightable3dgs,jiang2024gaussianshader,liang2024gsir}, while LE3D~\cite{jin2024le3d} uses a compact color network with per-Gaussian features and biases for RAW-domain rendering. Our work studies a Retinex-inspired RAW appearance representation with per-Gaussian base color, shared position- and view-conditioned modulation, and factor-specific regularization.

\paragraph{HDR and RAW-domain reconstruction.}
HDR-NeRF~\cite{huang2022hdr} reconstructs HDR radiance fields from LDR images captured at different exposures, while RawNeRF~\cite{mildenhall2022rawnerf} demonstrates reconstruction directly from noisy RAW observations. LE3D~\cite{jin2024le3d} extends RAW-domain reconstruction to real-time Gaussian rendering. From Chaos to Clarity~\cite{li2024chaos} combines Gaussian reconstruction with a learned noise extractor constrained by a sensor-noise prior, whereas HDRSplat~\cite{singh2024hdrsplat} uses PMRID denoising before Gaussian optimization. RetinexGS directly fits packed noisy RAW observations using channel-wise weights derived from DNG noise metadata. Camera projection is another relevant consideration: Fisheye-GS~\cite{liao2024fisheye} extends Gaussian projection to fisheye cameras, and From Chaos to Clarity~\cite{li2024chaos} maps rendered predictions into distorted image coordinates. Our implementation incorporates lens distortion into projected Gaussian centers and covariance footprints, enabling supervision on the packed RAW grid without spatially resampling the targets.

\paragraph{Retinex and low-light reconstruction.}
Single-image low-light enhancement methods use Retinex-based decomposition or learned enhancement models~\cite{guo2017lime,wei2018retinexnet,zhang2019kind,cai2023retinexformer,guo2020zerodce,wu2022uretinex,zheng2022sgz}. Applying such methods independently to individual views does not explicitly enforce multi-view consistency. Low-light neural rendering methods incorporate decomposition, illumination modeling, and physical priors into scene reconstruction~\cite{wang2023llnerf,cui2024alethnerf,zhou2025lita}. LL-Gaussian~\cite{sun2025ll} models reflectance and illumination attributes on Gaussian primitives for low-light sRGB reconstruction and enhancement, while DTGS~\cite{ma2025beyond} incorporates thermal supervision. RetinexGS focuses on factorized Gaussian appearance learned directly from noisy linear-RAW observations, combining separate appearance priors with sensor-informed supervision.

\paragraph{Geometric priors and initialization.}
Standard 3DGS uses an SfM point cloud for initialization~\cite{schoenberger2016sfm}, whose coverage can be limited in low-light and texture-poor regions. Sparse-view Gaussian methods use monocular depth priors to guide geometric optimization~\cite{xiong2024sparsegs,zhu2024fsgs,li2024dngaussian}. Learned reconstruction methods such as DUSt3R, MASt3R, and Fast3R provide dense geometric estimates~\cite{wang2024dust3r,leroy2024mast3r,yang2025fast3r}, and LL-Gaussian~\cite{sun2025ll} uses learned geometric priors for low-light initialization. Our initialization selectively supplements the SfM point cloud with depth-guided samples in dark regions with insufficient sparse-point coverage.

\section{Proposed Method}
\label{sec:method}

\begin{figure*}[t]
    \centering
    \includegraphics[width=\textwidth]{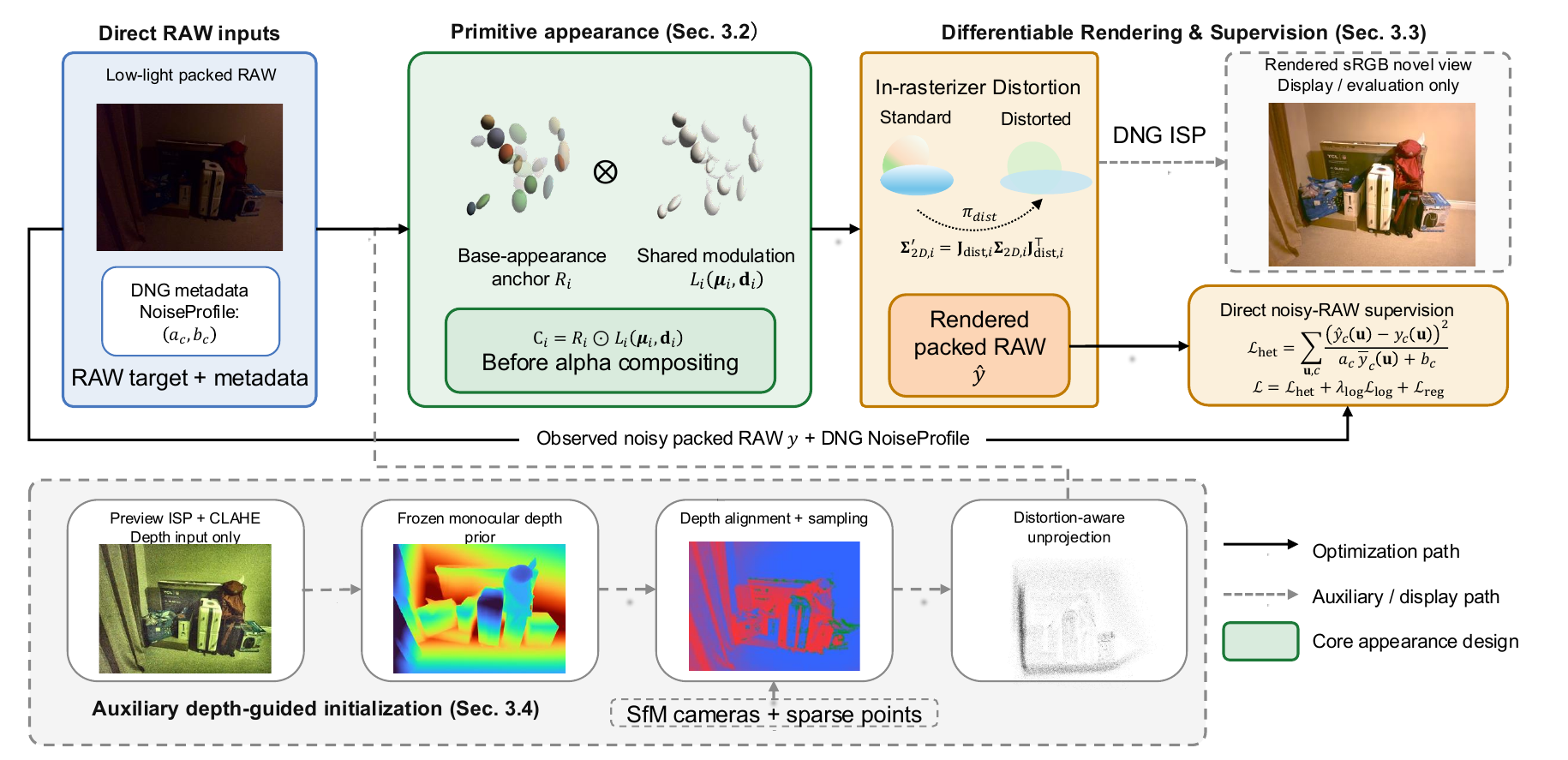}
    \caption{
    \textbf{RetinexGS overview.}
    Given noisy low-light RAW images and their DNG metadata, RetinexGS optimizes a Gaussian scene representation through three components.
    \textbf{(1)} A per-Gaussian base color $R_i$ and a shared, position- and view-conditioned modulation $L_i(\boldsymbol{\mu}_i,\mathbf{d}_i)$ define the primitive color $C_i=R_i\odot L_i(\boldsymbol{\mu}_i,\mathbf{d}_i)$. Separate regularization constrains the two factors.
    \textbf{(2)} Distortion-aware rasterization renders these colors directly onto the packed RAW grid, where training combines DNG-informed noise weighting, a log-domain reconstruction loss, and appearance regularization.
    \textbf{(3)} A frozen monocular depth prior supplies additional initialization points through dark-region sampling and distortion-aware unprojection.
    The DNG ISP branch is used only for visualization and evaluation.
    }
    \label{fig:pipeline}
\end{figure*}

RetinexGS (Fig.~\ref{fig:pipeline}) combines three components.
\textbf{(1) Factorized Gaussian appearance} (Sec.~\ref{sec:retinex}): each Gaussian stores a base color $R_i$ and receives a modulation $L_i$ predicted by a scene-shared network. The two factors are regularized separately and multiplied before alpha compositing.
\textbf{(2) Sensor-informed RAW supervision} (Sec.~\ref{sec:nll}): DNG noise parameters determine channel-wise, signal-dependent reconstruction weights, while distortion-aware rasterization enables supervision on the packed RAW grid without pre-undistorting the targets.
\textbf{(3) Auxiliary depth-guided initialization} (Sec.~\ref{sec:init}): monocular-depth estimates supplement the SfM point cloud in dark regions with insufficient sparse-point coverage.

\subsection{Preliminaries}
\label{sec:prelim}

\paragraph{3D Gaussian Splatting.}
3DGS represents a scene using anisotropic Gaussian primitives and renders them through front-to-back alpha compositing. Omitting the background contribution for clarity, the rendered color at pixel $\mathbf{u}$ is
\begin{equation}
C(\mathbf{u}) =
    \sum_i T_i(\mathbf{u})\,\alpha_i(\mathbf{u})\,C_i,
    \qquad
    T_i(\mathbf{u}) = \prod_{j<i}(1-\alpha_j(\mathbf{u})),
    \label{eq:alpha-compose}
\end{equation}
where primitives are ordered from front to back, $C_i\in\mathbb{R}^3$ is the color of primitive $i$, $\alpha_i(\mathbf{u})$ is its opacity contribution at the pixel, and $T_i(\mathbf{u})$ is the transmittance through the preceding primitives.

\paragraph{RAW noise model.}
We use the Gaussian approximation of the Poisson--Gaussian sensor noise model~\cite{foi2008practical}. For channel $k$, the observed RAW value follows $y_k\sim\mathcal{N}(\mu_k,a_k\mu_k+b_k)$, where $\mu_k$ is the underlying noise-free signal. The channel-wise parameters $a_k$ and $b_k$ describe signal-dependent and signal-independent noise variance, respectively, and are derived from DNG metadata.

\subsection{Retinex-Inspired Gaussian Appearance}
\label{sec:retinex}

\paragraph{Appearance factorization.}
RetinexGS represents each Gaussian's linear-RAW color as the product of a view-independent base appearance $R_i$ and a position- and view-dependent modulation $L_i$. Both factors are positive three-channel vectors. We parameterize the base appearance as
\begin{equation}
    R_i =
    \exp\!\left(
        \operatorname{clamp}_{(-\infty,3]}(c_i)
    \right),
    \label{eq:anchor}
\end{equation}
where $c_i\in\mathbb{R}^3$ is a trainable per-Gaussian log-color. For initialization, the SfM-derived RAW color is mapped into this log-color parameterization. The modulation and resulting primitive color are
\begin{equation}
\begin{aligned}
    L_i(\boldsymbol\mu_i,\mathbf d_i)
    &=
    \operatorname{clamp}_{[0.05,10]}\!\left[
        \operatorname{softplus}\!\left(
            \phi_L\!\left(
                \gamma_x(\boldsymbol\mu_i),
                \gamma_d(\mathbf d_i)
            \right)
        \right)
    \right],\\
    C_i
    &=
    R_i\odot L_i(\boldsymbol\mu_i,\mathbf d_i).
\end{aligned}
\label{eq:ret-param}
\end{equation}
Here, $\boldsymbol\mu_i$ is the Gaussian center, $\mathbf d_i$ is its unit viewing direction, and $\gamma_x$ and $\gamma_d$ are sinusoidal encodings. The network $\phi_L$ produces a three-channel output and is shared across all Gaussians in a scene. Thus, each primitive retains independently optimized base appearance, while modulation is coupled through a shared network. The output bias is initialized so that $L_i\approx\mathbf{1}$, allowing optimization to begin near the SfM-derived base appearance. Network details are provided in Appendix~\ref{app:arc}. We omit the arguments of $L_i$ below when they are unambiguous.

\paragraph{Composition before alpha blending.}
We form $R_i\odot L_i$ on each Gaussian before applying Eq.~\eqref{eq:alpha-compose}. An alternative is to rasterize the two factors separately and multiply the resulting image-space maps. The two operations are generally different:
\begin{proposition}[Non-commutativity of multiplication and alpha compositing]
\label{prop:noncommute}
Let $w_i=T_i\alpha_i$ denote the compositing weight of Gaussian $i$ at a pixel. Define
\[
\begin{aligned}
    C_{\mathrm{prim}}
    &= \sum_i w_i(R_i\odot L_i),\\
    C_{\mathrm{img}}
    &= \left(\sum_i w_iR_i\right)
       \odot
       \left(\sum_i w_iL_i\right).
\end{aligned}
\]
In general, $C_{\mathrm{prim}}\neq C_{\mathrm{img}}$.
\end{proposition}
The derivation and equality conditions are given in Appendix~\ref{app:proof-noncommute}. Primitive-level composition produces a single three-channel color attribute per Gaussian for rasterization. We compare its reconstruction quality and training cost with image-space composition under matched settings in Sec.~\ref{sec:composition-efficiency}.

\paragraph{Factor-specific regularization.}
The multiplicative representation admits a scale ambiguity: for $t>0$, replacing $(R_i,L_i)$ with $(tR_i,L_i/t)$ leaves $C_i$ unchanged whenever the rescaled factors remain within their bounds. We discourage scale drift and local fluctuations through
\begin{equation}
    \mathcal{L}_{\mathrm{reg}}
    =
    \lambda_{\mathrm{anc}}\mathcal{L}_{\mathrm{anc}}
    +
    \lambda_R\mathcal{L}_{\mathrm{smooth}}^{R}
    +
    \lambda_L\mathcal{L}_{\mathrm{smooth}}^{L}
    +
    \lambda_{\mathrm{chr}}
    \mathcal{L}_{\mathrm{smooth}}^{\mathrm{chr}},
    \label{eq:ret-prior}
\end{equation}
where $\mathcal{L}_{\mathrm{anc}}$ encourages the average modulation magnitude to remain near unity. The remaining terms encourage local consistency of log-base appearance, log-modulation magnitude, and modulation chrominance over 3D neighborhoods. These priors constrain the two appearance factors separately during optimization. Their definitions and weights are provided in Appendix~\ref{app:impl}.

\subsection{DNG-Informed Heteroscedastic Supervision}
\label{sec:nll}

RetinexGS fits noisy packed RAW observations using channel-wise sensor noise information. We derive reconstruction weights from DNG metadata, account for RAW packing and exposure normalization, and handle lens distortion during rendering to avoid spatially resampling the supervision targets.

\paragraph{Heteroscedastic weighting.}
We use a Gaussian approximation of the Poisson--Gaussian sensor noise model~\cite{foi2008practical}:
\begin{equation}
    y_c \sim \mathcal{N}(\mu_c,\; a_c\mu_c+b_c),
    \label{eq:noise-model-channel}
\end{equation}
where $c$ indexes the color channel, $y_c$ is the observed RAW value, and $\mu_c$ is the underlying noise-free signal. The parameters $(a_c,b_c)$ are obtained from the DNG \texttt{NoiseProfile} and expressed on the same intensity scale as the RAW observations.

Our three-channel packing averages the two Bayer green photosites in each $2{\times}2$ block. Assuming independent noise with shared green-channel parameters, the average follows the same variance model with
\[
    a_G=\frac{a_{\mathrm{green}}}{2},
    \qquad
    b_G=\frac{b_{\mathrm{green}}}{2}.
\]
We then account for the per-image exposure scale $e>0$. Under exposure normalization, the observations and noise parameters transform as
\[
    y'_c=\frac{y_c}{e},
    \qquad
    a'_c=\frac{a_c}{e},
    \qquad
    b'_c=\frac{b_c}{e^2}.
\]
This follows from the noise model under linear rescaling (Lemma~\ref{lem:exposure}, Appendix~\ref{app:impl}). Below, we drop the primes and use $y_c$, $a_c$, and $b_c$ for the exposure-normalized packed observations and their corresponding noise parameters.

Since $\mu_c$ is unknown, we obtain a nonnegative local signal estimate from the target:
\begin{equation}
    \bar{y}_c(\mathbf{u})
    =
    \mathrm{AvgPool}_{5\times5}(|y_c|)(\mathbf{u}).
\end{equation}
The absolute value accommodates negative observations produced by black-level subtraction. This estimate depends only on the target and remains fixed with respect to the optimized scene parameters. Denoting the rendered RAW prediction by $\hat{y}_c$, we minimize
\begin{equation}
    \mathcal{L}_{\mathrm{het}}
    =
    \sum_{\mathbf{u},c}
    \frac{
        \bigl(\hat{y}_c(\mathbf{u})-y_c(\mathbf{u})\bigr)^2
    }{
        a_c\bar{y}_c(\mathbf{u})+b_c
    },
    \label{eq:het}
\end{equation}
which is a DNG-informed weighted least-squares objective. Computing the variance estimates from the target keeps the reconstruction weights independent of the current model prediction.

We additionally use a log-domain L1 term to balance reconstruction gradients across the RAW dynamic range:
\begin{align}
    \mathcal{L}_{\log}
    &=
    \bigl\|
        \log(|\hat{y}|+\epsilon)
        -
        \log(|y|+\epsilon)
    \bigr\|_1,
    \qquad \epsilon=10^{-3},\\
    \mathcal{L}
    &=
    \mathcal{L}_{\mathrm{het}}
    +
    \lambda_{\log}\mathcal{L}_{\log}
    +
    \mathcal{L}_{\mathrm{reg}},
    \label{eq:objective}
\end{align}
where $\lambda_{\log}=0.4$ for the first 2,000 iterations and $0.8$ thereafter.

\paragraph{Avoiding target resampling.}
Eq.~\eqref{eq:het} uses a diagonal per-pixel noise approximation. Pre-undistorting RAW targets through spatial interpolation changes their noise variances and introduces correlations between neighboring pixels. We therefore retain the targets on the native packed RAW grid and incorporate lens distortion into the rasterizer. The rasterizer applies the OpenCV lens-distortion model to projected Gaussian centers and transforms their covariance footprints using the local distortion Jacobian. This aligns rendered predictions with the observed image coordinates without spatially resampling the RAW targets. Full projection equations are provided in Appendix~\ref{app:impl}.

\subsection{Depth-Guided Initialization for Low-Light Scenes}
\label{sec:init}

Low-light images often provide insufficient feature correspondences in dark and texture-poor regions, leaving gaps in the sparse SfM initialization. We supplement these regions with points derived from a monocular depth prior~\cite{Bochkovskii2024depth,yang2024depth}.

We apply CLAHE contrast enhancement to the images used for depth prediction, while keeping the RAW supervision targets unchanged. The predicted depths are aligned to sparse SfM depths through robust scale-and-shift fitting. We then sample additional points preferentially from dark regions with limited sparse-point coverage:
\begin{equation}
    p(\mathbf{u}) \propto
    \frac{1}{\bar{I}(\mathbf{u})+\epsilon}
    \cdot \mathbf{1}[\mathbf{u} \notin \mathcal{M}_{\mathrm{SfM}}],
    \label{eq:sampling}
\end{equation}
where $\bar{I}(\mathbf{u})$ is a locally smoothed intensity map and $\mathcal{M}_{\mathrm{SfM}}$ is a dilated mask of sparse SfM points projected into the current view. Sampled pixels are unprojected using their aligned depths and the same camera distortion model used for rendering. The resulting points augment the original SfM point cloud, improving initialization coverage in dark regions.



\section{Experiments}
\label{sec: exp}

\subsection{Experimental Setup}
\label{sec:exp-setup}
\label{sec:exp-impl} 

\paragraph{Datasets.}
We evaluate on the RawNeRF dataset~\cite{mildenhall2022rawnerf}, captured with a handheld iPhone X. Following RawNeRF and LE3D~\cite{jin2024le3d}, we use the three designated test scenes, \textit{pianotest}, \textit{officetest}, and \textit{yuccatest}, each with a tripod-stabilized clean reference, for quantitative evaluation, and six outdoor scenes without clean references (e.g., \textit{nightstreet}, \textit{gardenlights}, and \textit{parkstatue}) for qualitative evaluation. We additionally capture two low-light scenes with an iPhone 14 Pro Max at ISO 2000: \textit{housecorner}, an indoor nighttime scene with three tripod-stabilized reference views, and \textit{miningtunnel}, an underground passageway with sparse artificial lighting (Appendix~\ref{app:datacoll}).

\paragraph{Protocols and metrics.}
We evaluate under two protocols. In the \emph{native-pipeline} comparison, each method retains its own RAW processing and denoising pipeline (RetinexGS uses Bayer-cell half-packing), and sRGB reconstructions are compared against common held-out references. In the \emph{matched-input} comparison, RetinexGS, LE3D, and RawNeRF are retrained on identical three-channel half-packed RAW tensors at the same resolution, which standardizes input formation while retaining each method's appearance model and optimization procedure. HDRSplat+PMRID is evaluated only under the native-pipeline protocol, since its frozen denoiser defines its input interface. We report sRGB PSNR, SSIM~\cite{wang2004ssim}, and LPIPS~\cite{zhang2018lpips}, computed after per-channel affine alignment in each method's training domain and conversion with the RawNeRF post-processing pipeline (Appendix~\ref{app:data}). The matched-input comparison additionally reports linear-RAW metrics against identically normalized half-packed references, without affine fitting or ISP processing. Metrics for \textit{housecorner} are averaged over its three reference views.

\paragraph{Baselines.}
On RawNeRF, baseline results are taken from LE3D's unified comparison~\cite{jin2024le3d}, and we additionally reproduce HDRSplat with its released frozen PMRID checkpoint~\cite{singh2024hdrsplat} (Appendix~\ref{app:hdrsplat}). On our scenes, we compare against LDR-3DGS, trained with the standard 3DGS photometric loss on sRGB images from a fixed DNG-metadata ISP; LE3D; HDRSplat+PMRID; and Raw3DGS, our vanilla linear-RAW 3DGS with SH appearance, log-L1, and RawNeRF's weighted-L2 supervision~\cite{mildenhall2022rawnerf}, trained on the same half-packed inputs as RetinexGS.\footnote{Raw3DGS is our implementation and is distinct from RawGS, the SH-based RAW baseline reported by LE3D.} Per-scene Raw3DGS results on all five quantitative scenes are given in Appendix~\ref{app:raw3dgs}.

\paragraph{Implementation.}
RetinexGS is trained for 15{,}000 iterations with Adam~\cite{kingma2015adam}, with densification ending at iteration 7{,}000. The shared modulation network contains 9{,}379 parameters. Loss weights and schedules are given in Appendix~\ref{app:impl}.

\subsection{Native-Pipeline Comparison}
\label{sec:main-results}

\begin{table}[t]
    \centering
    \small
    \setlength{\tabcolsep}{6pt}
    \begin{tabular}{lcccc}
        \toprule
        Method & Iterations & PSNR$\uparrow$ & SSIM$\uparrow$ & LPIPS$\downarrow$ \\
        \midrule
        LDR-NeRF~\cite{mildenhall2020nerf}
        & -- & 20.04 & 0.5541 & 0.5669 \\
        LDR-3DGS~\cite{kerbl20233dgs}
        & 30k & 20.29 & 0.5477 & 0.5344 \\
        HDR-3DGS~\cite{jin2024le3d}
        & 30k & 20.33 & 0.5286 & 0.6563 \\
        RawNeRF~\cite{mildenhall2022rawnerf}
        & -- & 24.08 & \textbf{0.6100} & \underline{0.4952} \\
        RawGS~\cite{jin2024le3d}
        & 30k & 23.35 & 0.5843 & 0.5472 \\
        LE3D~\cite{jin2024le3d}
        & 30k & \underline{24.70} & \underline{0.6076} & 0.5071 \\
        HDRSplat+PMRID~\cite{singh2024hdrsplat}
        & 30k & 22.18 & 0.5160 & 0.5343 \\
        \midrule
        \textbf{RetinexGS}
        & 15k & \textbf{25.11} & 0.6061 & \textbf{0.4701} \\
        \bottomrule
    \end{tabular}
    \caption{\textbf{Native-pipeline comparison on RawNeRF.} Mean sRGB metrics over the three test scenes. Baseline values except HDRSplat+PMRID are taken from LE3D's unified comparison~\cite{jin2024le3d}; HDRSplat+PMRID is reproduced with its native preprocessing (Appendix~\ref{app:hdrsplat}). Bold and underline indicate the best and second-best values.}
    \label{tab:rawgs_compare}
\end{table}

\begin{table}[t]
    \centering
    \small
    \setlength{\tabcolsep}{5pt}
    \begin{tabular}{llcccc}
        \toprule
        Scene & Method & Iterations & PSNR$\uparrow$ & SSIM$\uparrow$ & LPIPS$\downarrow$ \\
        \midrule
        \multirow{5}{*}{\textit{miningtunnel}}
        & LDR-3DGS~\cite{kerbl20233dgs}
        & 30k & 17.29 & 0.3878 & 0.6209 \\
        & Raw3DGS (our impl.)
        & 30k & 11.85 & 0.3415 & 0.7327 \\
        & LE3D~\cite{jin2024le3d}
        & 30k & \underline{21.27} & \underline{0.4250} & \underline{0.6017} \\
        & HDRSplat+PMRID~\cite{singh2024hdrsplat}
        & 30k & 15.78 & 0.3340 & 0.7330 \\
        & \textbf{RetinexGS}
        & 15k & \textbf{23.47} & \textbf{0.4897} & \textbf{0.5053} \\
        \midrule
        \multirow{5}{*}{\textit{housecorner}}
        & LDR-3DGS~\cite{kerbl20233dgs}
        & 30k & 18.95 & 0.6051 & 0.6264 \\
        & Raw3DGS (our impl.)
        & 30k & \underline{24.61} & \underline{0.6255} & \underline{0.5222} \\
        & LE3D~\cite{jin2024le3d}
        & 30k & 22.74 & 0.6224 & 0.5610 \\
        & HDRSplat+PMRID~\cite{singh2024hdrsplat}
        & 30k & 16.57 & 0.5170 & 0.7550 \\
        & \textbf{RetinexGS}
        & 15k & \textbf{26.27} & \textbf{0.6919} & \textbf{0.4401} \\
        \bottomrule
    \end{tabular}
    \caption{\textbf{Native-pipeline comparison on self-collected scenes.} All methods are trained by us. RawNeRF on these scenes is evaluated under the matched-input protocol (Table~\ref{tab:matched}). Bold and underline indicate the best and second-best values.}
    \label{tab:ours_compare}
\end{table}

Tables~\ref{tab:rawgs_compare} and~\ref{tab:ours_compare} report native-pipeline results on the public and self-collected scenes. On the RawNeRF test scenes, RetinexGS achieves the highest mean PSNR and lowest LPIPS, with SSIM within 0.004 of the best baseline. Compared with LE3D, it improves PSNR by 0.41\,dB and reduces LPIPS by 0.037 using half the training iterations. On \textit{miningtunnel} and \textit{housecorner}, RetinexGS is best on all three metrics, exceeding the strongest baseline in PSNR by 2.20\,dB (LE3D) and 1.66\,dB (Raw3DGS), respectively. It also improves all three metrics over HDRSplat+PMRID on each of the five quantitative scenes (Table~\ref{tab:hdrsplat-per-scene}).

\begin{figure*}[t]
    \centering
    \includegraphics[width=\textwidth]{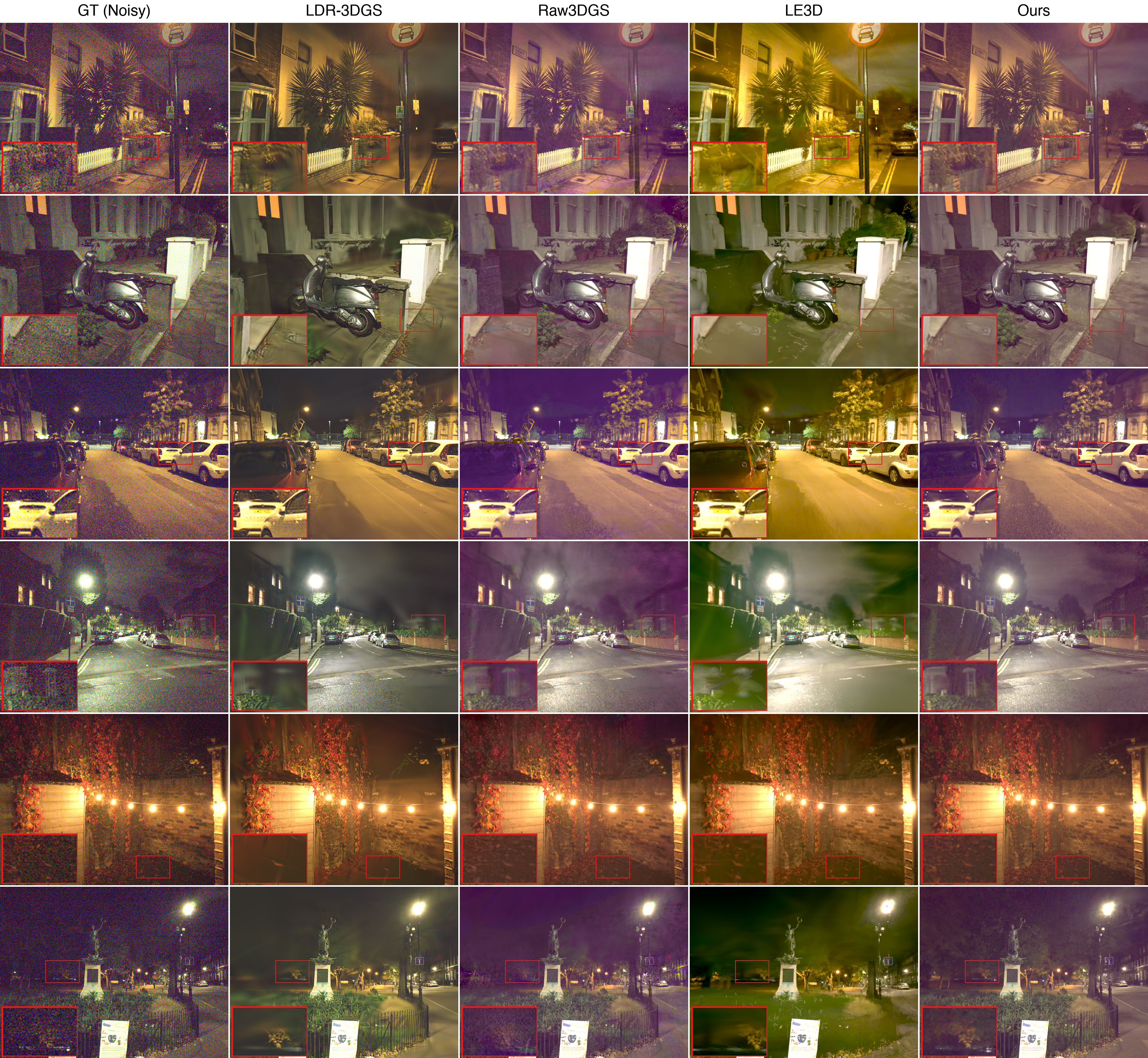}
    \caption{\textbf{Qualitative comparison on RawNeRF scenes.} Held-out novel views, shown without per-channel affine correction since clean references are unavailable at these poses. Compared with RawGS, LE3D, and LDR-3DGS, RetinexGS produces fewer floating artifacts in dark regions and preserves finer details.}
    \label{fig:main-comparison}
\end{figure*}

Fig.~\ref{fig:main-comparison} shows held-out novel views on the qualitative RawNeRF scenes, using the LLFF one-in-eight holdout split~\cite{mildenhall2019local} adopted by RawNeRF and LE3D. RetinexGS preserves fine details and dark-region structure, with fewer floating artifacts and less visible color distortion than the baselines. Renderings on the self-collected scenes are shown in Fig.~\ref{fig:our-comparison}.

\subsection{Matched-Input Comparison}
\label{sec:matched-input}

\begin{table}[t]
    \centering
    \small
    \setlength{\tabcolsep}{5pt}
    \begin{tabular}{lccccc}
        \toprule
        \multirow{2}{*}{Method}
        & \multicolumn{3}{c}{sRGB}
        & \multicolumn{2}{c}{Linear RAW} \\
        \cmidrule(lr){2-4} \cmidrule(lr){5-6}
        & PSNR$\uparrow$ & SSIM$\uparrow$ & LPIPS$\downarrow$
        & PSNR$\uparrow$ & SSIM$\uparrow$ \\
        \midrule
        RawNeRF~\cite{mildenhall2022rawnerf}
        & 22.31 & 0.484 & 0.550 & 60.92 & 0.9977 \\
        LE3D~\cite{jin2024le3d}
        & 22.83 & 0.550 & 0.553 & 61.14 & 0.9977 \\
        \midrule
        \textbf{RetinexGS}
        & \textbf{25.01} & \textbf{0.600} & \textbf{0.471} & \textbf{63.39} & \textbf{0.9987} \\
        \bottomrule
    \end{tabular}
    \caption{\textbf{Matched-input comparison.} Five-scene means for methods retrained on identical half-packed RAW tensors at the same resolution. sRGB metrics use per-channel affine alignment in linear RAW before ISP conversion; linear-RAW metrics use neither. RetinexGS attains the highest PSNR on every scene in both domains; per-scene values are given in Table~\ref{tab:matched-full}. Best values are shown in bold.}
    \label{tab:matched}
    \label{tab:matched-psnr} 
\end{table}

Table~\ref{tab:matched} compares methods retrained on the same half-packed RAW inputs as RetinexGS, with training configurations given in Appendix~\ref{app:matched-full}. RetinexGS attains the highest PSNR on every scene in both sRGB and linear RAW. Its mean sRGB PSNR exceeds LE3D and RawNeRF by 2.18 and 2.71\,dB, with corresponding linear-RAW gains of 2.25 and 2.47\,dB, and it also achieves the best mean SSIM and LPIPS. Because the linear-RAW metrics involve no affine fitting or ISP conversion, the ranking does not depend on the sRGB evaluation pipeline. Raw3DGS, trained on the same inputs, reaches a mean sRGB PSNR of 19.21\,dB after 30k iterations (Appendix~\ref{app:raw3dgs}; Fig.~\ref{fig:teaser}).

\subsection{Component Ablations}
\label{sec:component-ablation}

\begin{table}[t]
    \centering
    \small
    \setlength{\tabcolsep}{4pt}
    \begin{tabular}{lcccccc}
        \toprule
        \multirow{2}{*}{Variant}
        & \multicolumn{3}{c}{\textit{pianotest}}
        & \multicolumn{3}{c}{\textit{miningtunnel}} \\
        \cmidrule(lr){2-4} \cmidrule(lr){5-7}
        & PSNR$\uparrow$ & SSIM$\uparrow$ & LPIPS$\downarrow$
        & PSNR$\uparrow$ & SSIM$\uparrow$ & LPIPS$\downarrow$ \\
        \midrule
        SH appearance
        & 19.20 & 0.4651 & 0.5829 & 19.53 & 0.4019 & 0.6322 \\
        RawNeRF weighting~\cite{mildenhall2022rawnerf}
        & 25.59 & \textbf{0.6238} & 0.4800 & \textbf{23.47} & 0.4878 & 0.5117 \\
        Log-L1 only
        & 25.42 & 0.6168 & 0.4801 & 22.19 & 0.4566 & 0.5485 \\
        w/o factor reg.
        & 25.77 & 0.6213 & 0.4630 & 23.27 & 0.4894 & 0.5107 \\
        w/o depth init.
        & 25.80 & 0.6209 & 0.4725 & 22.81 & 0.4477 & 0.5837 \\
        \midrule
        \textbf{Full model}
        & \textbf{25.84} & 0.6230 & \textbf{0.4610} & \textbf{23.47} & \textbf{0.4897} & \textbf{0.5053} \\
        \bottomrule
    \end{tabular}
    \caption{\textbf{Component ablations.} sRGB metrics. RawNeRF weighting replaces the DNG-informed weighting; log-L1 only removes the heteroscedastic term. Best values are shown in bold.}
    \label{tab:ablation}
\end{table}

Table~\ref{tab:ablation} ablates individual components on \textit{pianotest}, an indoor low-light scene, and \textit{miningtunnel}, a sparsely illuminated underground scene.

\paragraph{Appearance parameterization.}
Replacing the factorized appearance with standard SH reduces PSNR by 6.64 and 3.94\,dB on the two scenes. We further evaluate LE3D's appearance parameterization within the same training framework across five scenes, using 15k iterations, seed 0, and LE3D's native appearance learning rate. RetinexGS improves mean sRGB PSNR by 0.34\,dB, achieving higher PSNR and SSIM and lower LPIPS on all five scenes.

\paragraph{Noise supervision.}
Replacing the DNG-informed weighting with RawNeRF's weighted-L2 loss~\cite{mildenhall2022rawnerf} yields similar PSNR and SSIM but higher LPIPS on both scenes, and introduces a visible purple cast in low-signal regions (Appendix~\ref{app:nll_vs_l2}). Using log-L1 alone degrades all metrics on both scenes, most notably on \textit{miningtunnel}, where PSNR drops by 1.28\,dB.

\paragraph{Regularization and initialization.}
Factor-specific regularization provides modest gains on both scenes. Depth-guided initialization matters more on \textit{miningtunnel}, where removing it lowers PSNR by 0.66\,dB and raises LPIPS by 0.078, consistent with its role in supplementing sparse SfM coverage in dark regions.

\subsection{Composition Placement and Efficiency}
\label{sec:composition-efficiency}
\label{sec:efficiency} 

\paragraph{Composition placement.}
Under identical half-packed inputs, training views, and 15k-iteration budgets, primitive-level and image-space composition reach nearly identical five-scene quality: 25.01 vs.\ 24.96\,dB sRGB PSNR and 63.39 vs.\ 63.20\,dB linear-RAW PSNR (per-scene results in Appendix~\ref{app:placement-full}). Although the two operators are not equivalent (Proposition~\ref{prop:noncommute}), joint optimization of the appearance factors largely compensates for the difference. Image-space composition, however, rasterizes and backpropagates through two three-channel factor maps, whereas primitive-level composition rasterizes their single product. In our implementation, this approximately halves wall-clock training time, which motivates our choice. The full model trains in 0.4--0.6 hours per scene on a single NVIDIA RTX 4090.

\paragraph{Rendering and storage.}
At $2016\times1512$ resolution on an RTX 4090, RetinexGS renders at 91--145\,FPS. This is below a matched SH control with identical geometry and rasterization (148--215\,FPS), reflecting the cost of evaluating the shared modulation network, while peak memory remains comparable or lower (Table~\ref{tab:efficiency} in Appendix~\ref{app:efficiency}). The modulation network occupies approximately 37\,KB in FP32, and complete scene models occupy 14--34\,MB.

\section{Conclusion}
\label{sec:conclusion}

We presented \emph{RetinexGS}, a RAW-domain Gaussian Splatting framework for low-light novel-view synthesis. Its appearance model combines a per-Gaussian base factor with a compact scene-shared modulation field, enabling factor-specific regularization and composition before alpha blending. DNG-informed noise weighting and distortion-aware rendering support direct optimization against noisy RAW observations, while depth-guided initialization supplements sparse geometry in dark regions. Experiments on public and self-collected scenes demonstrate improved reconstruction quality under both native-pipeline and matched-input comparisons. Primitive-level composition achieves quality comparable to image-space composition while approximately halving training time in our implementation. These results support combining structured Gaussian appearance with sensor-informed optimization for low-light RAW reconstruction.

\paragraph{Limitations and future work.}
RetinexGS currently targets static scenes and uses DNG noise metadata for supervision. Residual floating artifacts can remain in extremely low-signal, texture-poor regions. Future work includes evaluation across a broader range of sensors, capture-adaptive noise estimation, and stronger geometric constraints for poorly observed regions. Extending the learned appearance factors to support relighting and appearance editing is another promising direction.

\bibliographystyle{unsrtnat}
\bibliography{reference}


\newpage
\appendix

\section{Architecture Details}
\label{app:arc}

The evaluated appearance path contains one per-Gaussian base anchor and one scene-shared modulation network (Table~\ref{tab:appearance-architecture}).

\begin{table}[h]
\centering
\small
\begin{tabular}{lll}
\toprule
Quantity & Input / storage & Output transform \\
\midrule
$R_i$ & learnable $c_i\in\mathbb R^3$ per Gaussian
      & $\exp(\operatorname{clamp}_{(-\infty,3]}(c_i))$ \\
$L_i$ & encoded $\boldsymbol\mu_i$ and $\mathbf d_i$; shared MLP
      & $\operatorname{clamp}_{[0.05,10]}(\operatorname{softplus}(\cdot))$ \\
$C_i$ & $R_i$ and $L_i(\boldsymbol\mu_i,\mathbf d_i)$
      & $R_i\odot L_i$ \\
\bottomrule
\end{tabular}
\caption{Factor architecture shared by the two composition variants. The output bias initializes $L_i$ near one. The base anchor is directly optimized, and the shared modulation field is parameterized by $\phi_L$. The final row is the primitive-level composition used by RetinexGS; the image-space variant instead rasterizes $R_i$ and $L_i$ separately and multiplies the resulting maps.}
\label{tab:appearance-architecture}
\end{table}

The position branch of $\phi_L$ applies two linear layers with hidden width 64 to the position encoding. Its 64-dimensional output is concatenated with the view-direction encoding and passed through a two-linear-layer view branch with a 64-wide hidden layer and a three-channel output. All parameters of $\phi_L$ are scene-shared. Active parameter counts and checkpoint sizes include this network and the directly optimized base anchors.

\section{Proof of Proposition~\ref{prop:noncommute}}
\label{app:proof-noncommute}

\paragraph{Setup.}
Consider a fixed pixel covered by a set of Gaussian primitives. Let $R_i,L_i\in\mathbb{R}^{3}_{>0}$ denote the base and modulation factors of primitive $i$, and let $w_i=T_i\alpha_i\ge 0$ be its alpha-compositing weight at the pixel. In 3DGS alpha compositing, the accumulated opacity weight satisfies $W=\sum_i w_i\le 1$. We compare two ways of combining the multiplicative factors with alpha compositing.

The primitive-level composition used by RetinexGS first forms the appearance product for each primitive and then alpha-composites:
\begin{equation}
    C^{\mathrm{pp}}
    =
    \sum_i w_i (R_i \odot L_i).
    \label{eq:app-pp}
\end{equation}
In contrast, image-space composition separately composites the base and modulation maps and multiplies them after rasterization (superscript $\mathrm{pr}$ for post-rasterization):
\begin{equation}
    C^{\mathrm{pr}}
    =
    \left(\sum_i w_i R_i\right)
    \odot
    \left(\sum_j w_j L_j\right).
    \label{eq:app-pr}
\end{equation}

\paragraph{Non-commutativity.}
Expanding Eq.~\eqref{eq:app-pr} gives
\begin{equation}
    C^{\mathrm{pr}}
    =
    \sum_i w_i^2 (R_i \odot L_i)
    +
    \sum_{i\ne j} w_iw_j (R_i \odot L_j).
    \label{eq:app-pr-expand}
\end{equation}
Therefore, the difference between the two compositions is
\begin{equation}
    C^{\mathrm{pp}} - C^{\mathrm{pr}}
    =
    \sum_i (w_i-w_i^2)(R_i\odot L_i)
    -
    \sum_{i\ne j} w_iw_j(R_i\odot L_j).
    \label{eq:app-gap}
\end{equation}
The second term contains cross-products $R_i\odot L_j$ between different primitives, which are not part of the intended per-primitive appearance product. These cross-products arise solely because the base and modulation factors were separately alpha-composited before multiplication. Hence alpha compositing and element-wise factor multiplication do not commute in general.

\paragraph{Relation to covariance.}
For completeness, let $W=\sum_i w_i$ and define normalized weights $p_i=w_i/W$ when $W>0$. For each color channel $c$,
\begin{align}
    C^{\mathrm{pp}}_c
    &= W\,\mathbb{E}_{p}[R_cL_c],\\
    C^{\mathrm{pr}}_c
    &= W^2\,\mathbb{E}_{p}[R_c]\,\mathbb{E}_{p}[L_c].
\end{align}
Thus,
\begin{equation}
    C^{\mathrm{pp}}_c - C^{\mathrm{pr}}_c
    =
    W\,\mathrm{Cov}_{p}(R_c,L_c)
    +
    W(1-W)\,
    \mathbb{E}_{p}[R_c]\,
    \mathbb{E}_{p}[L_c].
    \label{eq:app-cov}
\end{equation}
When $W=1$, the discrepancy reduces to the weighted covariance between the base and modulation factors across overlapping primitives. For general alpha compositing, however, $W$ can be smaller than one, and the additional opacity term in Eq.~\eqref{eq:app-cov} is nonzero. This shows that post-rasterization multiplication can disagree with primitive-level composition even when the two factors are uncorrelated across primitives.

\paragraph{Equality cases.}
For $W>0$, equality $C^{\mathrm{pp}}=C^{\mathrm{pr}}$ holds if and only if, for every color channel $c$,
\begin{equation}
    \mathrm{Cov}_{p}(R_c,L_c)
    = (W-1)\,\mathbb{E}_{p}[R_c]\,\mathbb{E}_{p}[L_c].
    \label{eq:app-equality}
\end{equation}
This is the complete channel-wise condition, including the accumulated coverage $W=\sum_i w_i$. In particular, for the strictly positive factors used by RetinexGS, zero covariance is sufficient only under full coverage ($W=1$); at a semi-transparent boundary with $W<1$, uncorrelated factors alone do not give equality. When $W=0$, both renderings are trivially zero. With $W>0$, simple equality cases include fully opaque single-primitive coverage; more generally, equality requires the exact channel-wise cancellation in Eq.~\eqref{eq:app-equality}. In scenes with multiple overlapping semi-transparent Gaussians and varying base and modulation factors, the two compositions are not equivalent in general.

\paragraph{Interpretation.}
This result distinguishes the two computational graphs but does not predict which one will optimize to lower held-out error. Because all appearance parameters are learned jointly, either model can compensate for part of the operator difference. The matched experiments confirm this: despite non-commutativity, the two placements attain nearly identical NVS metrics. Our selection of primitive-level composition is therefore based on measured training efficiency rather than reconstruction accuracy.

\section{Implementation Details}
\label{app:impl}

\paragraph{Distortion-aware rasterizer.}
To avoid additional resampling-induced correlations in the diagonal per-pixel weighting model in Eq.~\ref{eq:het}, we move lens distortion into the forward rendering path rather than pre-undistorting RAW targets. For each Gaussian centered at $\boldsymbol{\mu}_i$, let $(\tilde{x}, \tilde{y}) = \pi(\boldsymbol{\mu}_i)$ denote the undistorted projected coordinates from the standard pinhole projection $\pi(\cdot)$. Given OpenCV distortion parameters $(k_1, k_2, p_1, p_2)$, the distorted coordinates follow the radial--tangential model:
\begin{equation}
    (x_d, y_d) = (1 + k_1 r^2 + k_2 r^4)(\tilde{x}, \tilde{y})
    + \mathbf{t}(\tilde{x}, \tilde{y}; p_1, p_2),
    \quad r^2 = \tilde{x}^2 + \tilde{y}^2,
    \label{eq:distort}
\end{equation}
where $\mathbf{t}$ denotes the tangential distortion term. Since distortion alters both the projected center and the local screen-space footprint, we define the local $2\times2$ distortion Jacobian
\begin{equation}
    \mathbf{J}_{\mathrm{dist},i}
    =
    \frac{\partial(x_d,y_d)}{\partial(\tilde{x},\tilde{y})}
    \big|_{\pi(\boldsymbol\mu_i)}.
\end{equation}
If $\boldsymbol\Sigma_{2D,i}$ is the standard undistorted screen-space covariance, the distorted footprint is
\begin{equation}
    \boldsymbol\Sigma'_{2D,i}
    =
    \mathbf{J}_{\mathrm{dist},i}
    \boldsymbol\Sigma_{2D,i}
    \mathbf{J}_{\mathrm{dist},i}^{\mathsf T}.
    \label{eq:distorted-covariance}
\end{equation}
Equivalently, with the standard $2\times3$ projection Jacobian $\mathbf{J}_{\pi,i}$, the composed projection Jacobian is $\mathbf{J}'_i=\mathbf{J}_{\mathrm{dist},i}\mathbf{J}_{\pi,i}$ and may be applied directly to the 3D covariance. Using $\mathbf{J}_{\pi,i}$ alone misestimates the distorted screen-space footprint of Gaussians near image boundaries. The projection is implemented as a fused CUDA kernel with analytical Jacobians.

\paragraph{Training.}
All RetinexGS variants are trained for 15{,}000 iterations on a single NVIDIA RTX 4090 GPU, with densification stopped at iteration 7{,}000. We use the Adam optimizer with learning rates following the standard 3DGS schedule~\cite{kerbl20233dgs}. The densification gradient threshold is increased from the 3DGS default $2 \times 10^{-4}$ to $1.5 \times 10^{-3}$ to suppress over-densification under noisy RAW supervision, keeping the final Gaussian count between $10^5$ and $5 \times 10^5$ across scenes.

\paragraph{Composition control.}
The primitive-level and image-space arms use identical factor parameterization, regularization, initialization, training views, optimization schedule, densification, and evaluation. The composition mode is logged for each run and asserted in the training loop before the first render, so the placement comparison changes only the rendering operator. The image-space arm accumulates and differentiates two three-channel factor maps, whereas the primitive-level arm renders their single three-channel product.

\paragraph{Noise model.}
The local signal level is estimated from the observed noisy target as
\[
\bar{y}(\mathbf{u}) = \mathrm{AvgPool}_{5 \times 5}(|y|).
\]
The absolute value handles negative values arising from black-level subtraction. $\bar{y}$ is computed once and held fixed (stop-gradient) during optimization of the heteroscedastic term (Eq.~\ref{eq:het}).

\paragraph{Depth-guided initialization.}
We use 40\% of training views (capped at 100) for depth augmentation, sampling $N = 3{,}000$ new points per view. CLAHE contrast enhancement (clip limit 2.0, tile grid $8\!\times\!8$) is applied to the input image before monocular depth prediction only; supervision images are unchanged. For the sampling distribution in Eq.~\ref{eq:sampling}, $\bar I$ is computed via Gaussian smoothing ($\sigma=5$ pixels) on the per-pixel intensity, $\epsilon=10^{-3}$ prevents divergence in extremely dark regions, and $\mathcal{M}_{\mathrm{COLMAP}}$ is dilated by 8 pixels to suppress oversampling near existing SfM points.

\paragraph{Retinex regularization.}
We apply four regularization terms on per-Gaussian Retinex attributes. We sample $|\mathcal{S}| = 2{,}048$ anchor Gaussians per iteration and find their $k = 5$ nearest neighbors $\mathcal{N}_k(i)$ in 3D position. The full regularization is:
\begin{equation}
    \mathcal{L}_{\mathrm{reg}} =
    \lambda_{\mathrm{anc}} \mathcal{L}_{\mathrm{anc}} +
    \lambda_R \mathcal{L}_{\mathrm{smooth}}^{R} +
    \lambda_L \mathcal{L}_{\mathrm{smooth}}^{L} +
    \lambda_{\mathrm{chr}} \mathcal{L}_{\mathrm{smooth}}^{\mathrm{chr}}.
\end{equation}

\textbf{Modulation anchor.} Per-Gaussian mean modulation is anchored near unity to choose a global scale gauge:
\begin{equation}
    \mathcal{L}_{\mathrm{anc}} =
    \frac{1}{N} \sum_i \Bigl( \tfrac{1}{3} \sum_c L_{i,c} - 1 \Bigr)^2.
\end{equation}

\textbf{Base-anchor smoothness.}
We smooth the log-color anchor $c_i$, the only learnable per-Gaussian appearance parameter and a direct proxy for $\log R_i$:
\begin{equation}
    \mathcal{L}_{\mathrm{smooth}}^{R} =
    \frac{1}{|\mathcal{S}|k}
    \sum_{i \in \mathcal{S}}
    \sum_{j \in \mathcal{N}_k(i)}
    \| c_i - c_j \|_2^2 .
\end{equation}

\textbf{Modulation-magnitude smoothness.} Log-modulation is smoothed with linear warm-up:
\begin{equation}
    \mathcal{L}_{\mathrm{smooth}}^{L} =
    w(t) \, \frac{1}{|\mathcal{S}|k}
    \sum_{i \in \mathcal{S}} \sum_{j \in \mathcal{N}_k(i)}
    \| \log L_i - \log L_j \|_2^2 ,
\end{equation}
where $w(t) = \min(1, (t - 2000)/3000)$ for $t > 2000$, zero otherwise.

\textbf{Chrominance smoothness.} Normalized modulation direction is smoothed to discourage the shared branch from absorbing view-independent color shifts:
\begin{equation}
    \mathcal{L}_{\mathrm{smooth}}^{\mathrm{chr}} =
    \frac{1}{|\mathcal{S}|k}
    \sum_{i \in \mathcal{S}} \sum_{j \in \mathcal{N}_k(i)}
    \Bigl\| \frac{L_i}{\bar{L}_i} - \frac{L_j}{\bar{L}_j} \Bigr\|_2^2 ,
\end{equation}
where $\bar{L}_i = \tfrac{1}{3} \sum_c L_{i,c}$.

\textbf{Loss schedule and weights.} Regularization terms activate progressively to allow geometry to stabilize before appearance constraints (Table~\ref{tab:loss-weights}).

\begin{table}[h]
\centering
\small
\begin{tabular}{lcc}
\toprule
Loss term & Weight & Active from \\
\midrule
$\lambda_{\mathrm{anc}}$ & 0.001 & iter 1000 \\
$\lambda_L$ (modulation magnitude) & 0.02 & iter 2000 (3k warm-up) \\
$\lambda_R$ (base anchor) & 0.005 & iter 4000 \\
$\lambda_{\mathrm{chr}}$ & 0.005 & iter 4000 \\
$\lambda_{\log}$ (iter $\le$ 2000) & 0.4 & iter 0 \\
$\lambda_{\log}$ (iter $>$ 2000) & 0.8 & iter 2000 \\
\bottomrule
\end{tabular}
\caption{Loss weights and activation schedule.}
\label{tab:loss-weights}
\end{table}

\paragraph{Exposure-rescaling closure.}
We formally state the exposure-rescaling property used in Sec.~\ref{sec:nll}.

\begin{lemma}[Exposure-rescaling closure]
\label{lem:exposure}
If $y \sim \mathcal{N}(\mu, a\mu + b)$ and $y' = y/e$ for scalar $e > 0$, then $y' \sim \mathcal{N}(\mu', a'\mu' + b')$ with $\mu' = \mu/e$, $a' = a/e$, $b' = b/e^2$.
\end{lemma}
\begin{proof}
$\mathbb{E}[y'] = \mu/e = \mu'$. $\mathrm{Var}(y') = \mathrm{Var}(y)/e^2 = (a\mu+b)/e^2 = (a/e)(\mu/e) + b/e^2 = a'\mu' + b'$.
\end{proof}

The DNG parameters $(a,b)$ are therefore rescaled to $(a/e,b/e^2)$ after exposure normalization. The same closure applies independently per channel, so the parameters used in Eq.~\ref{eq:het} remain on the same scale as the normalized observations.

\section{Dataset and Evaluation Details}
\label{app:data}

\paragraph{RawNeRF benchmark.}
We follow the standard train/test split of RawNeRF~\cite{mildenhall2022rawnerf}, using the three designated test scenes (\textit{pianotest}, \textit{officetest}, \textit{yuccatest}) for quantitative evaluation. In the native-pipeline comparison on these scenes, results for LDR-NeRF, LDR-3DGS, HDR-3DGS, RawNeRF, RawGS, and LE3D are taken from~\cite{jin2024le3d}; all other results, and all results on the self-collected scenes, are produced by us.

\paragraph{Evaluation pipeline.}
RAW predictions are converted to sRGB with the RawNeRF post-processing pipeline~\cite{mildenhall2022rawnerf}: white balance, a color matrix, 97th-percentile exposure normalization, and sRGB gamma. Per-channel affine alignment is fit in each method's training domain, i.e., linear RAW for RAW-based methods and sRGB for LDR-3DGS. Linear-RAW metrics in the matched-input comparison are computed against identically normalized half-packed references, without affine fitting or ISP processing.

\paragraph{Self-collected low-light RAW dataset.}
\label{app:datacoll}
We capture two challenging low-light scenes with an iPhone 14 Pro Max using the Halide camera app\footnote{\url{https://halide.cam/}} at ISO 2000 and 4K resolution. \textit{miningtunnel} is an underground passageway with sparse artificial lighting and large textureless regions; \textit{housecorner} is an indoor scene with warm artificial lighting. Each scene contains 100--110 training views captured handheld.

For evaluation, we provide one reference view for \textit{miningtunnel} and three for \textit{housecorner} at different camera positions, with metrics averaged across the three. The reference views use $2$--$3\times$ longer exposures than the training views. Each reference is generated by fixing the camera on a tripod and averaging approximately 100 consecutive RAW frames from the same viewpoint to produce a low-noise reference stored as a \texttt{.npy} array. All reference views are held out from training. Images are processed using the same exposure normalization as RawNeRF~\cite{mildenhall2022rawnerf}.

\section{HDRSplat Reproduction}
\label{app:hdrsplat}

We reproduce the paper-described HDRSplat pipeline~\cite{singh2024hdrsplat} using the released frozen PMRID Bayer-space denoiser and pretrained checkpoint, bilinear demosaicing, RAW-space supervision, degree-3 SH, the released rasterization and optimization settings, and 30k training iterations. In the released scene loader, \texttt{demosaic} is hard-coded into the positional \texttt{denoise\_method} argument, while the command-line PMRID option is passed to the following argument and therefore does not affect image loading. We apply a minimal argument-propagation fix and verify from the resolved input paths that \texttt{PMRID\_raw}, rather than the original demosaiced noisy frame, is passed to training. No pretrained weights, model architecture, loss, rasterizer, or optimization settings are otherwise changed.

Table~\ref{tab:hdrsplat-footprint} compares optimized primitive counts, and Table~\ref{tab:hdrsplat-per-scene} reports per-scene metrics under the common output-domain evaluation.

\begin{table}[h]
\centering
\small
\setlength{\tabcolsep}{5pt}
\begin{tabular}{lrrr}
\toprule
Scene & HDRSplat Gaussians & RetinexGS Gaussians & Reduction\\
\midrule
\textit{pianotest} & 505k & 340k & 33\%\\
\textit{officetest} & 579k & 273k & 53\%\\
\textit{yuccatest} & 199k & 133k & 33\%\\
\textit{miningtunnel} & 302k & 225k & 25\%\\
\textit{housecorner} & 460k & 138k & 70\%\\
\midrule
Mean & 409k & 222k & 46\%\\
\bottomrule
\end{tabular}
\caption{Optimized primitive counts for HDRSplat+PMRID (30k iterations, degree-3 SH) and RetinexGS (15k iterations). The mean reduction is computed from the mean counts.}
\label{tab:hdrsplat-footprint}
\end{table}

\begin{table}[h]
\centering
\scriptsize
\setlength{\tabcolsep}{4pt}
\begin{tabular}{lccc}
\toprule
Scene & HDRSplat+PMRID & RetinexGS & Difference\\
\midrule
\textit{pianotest}
& 22.84 / 0.565 / 0.490
& 25.84 / 0.623 / 0.461
& $+3.00$ / $+0.058$ / $-0.029$\\
\textit{officetest}
& 23.97 / 0.587 / 0.475
& 25.58 / 0.635 / 0.440
& $+1.61$ / $+0.048$ / $-0.035$\\
\textit{yuccatest}
& 19.73 / 0.396 / 0.638
& 23.91 / 0.560 / 0.509
& $+4.18$ / $+0.164$ / $-0.129$\\
\textit{miningtunnel}
& 15.78 / 0.334 / 0.733
& 23.47 / 0.490 / 0.505
& $+7.69$ / $+0.156$ / $-0.228$\\
\textit{housecorner}
& 16.57 / 0.517 / 0.755
& 26.27 / 0.692 / 0.440
& $+9.70$ / $+0.175$ / $-0.315$\\
\bottomrule
\end{tabular}
\caption{Per-scene direct comparison in sRGB PSNR / SSIM / LPIPS under the common output-domain evaluation. The public RawNeRF scenes support the primary comparison; the larger collected-scene gaps are interpreted only as end-to-end robustness results.}
\label{tab:hdrsplat-per-scene}
\end{table}

As a cross-check, a second evaluation implementation reproduces the \textit{miningtunnel} PSNR difference to within 0.02\,dB. We do not use the collected scenes to claim general cross-sensor behavior of PMRID.

\section{Five-Scene Raw3DGS Control}
\label{app:raw3dgs}

We train our vanilla Raw3DGS implementation on the same half-packed tensors used by RetinexGS. Raw3DGS retains SH appearance and uses log-L1 plus the weighted-L2 supervision of RawNeRF~\cite{mildenhall2022rawnerf}; each scene is optimized for $30{,}000$ iterations. Table~\ref{tab:raw3dgs-perscene} reports the values under the same affine-sRGB evaluation used elsewhere in the paper.

\begin{table}[h]
\centering
\small
\setlength{\tabcolsep}{6pt}
\begin{tabular}{lcccc}
\toprule
Scene & Iter. & PSNR$\uparrow$ & SSIM$\uparrow$ & LPIPS$\downarrow$ \\
\midrule
\textit{pianotest}    & 30k & 19.23 & 0.5675 & 0.6252 \\
\textit{officetest}   & 30k & 21.85 & 0.5932 & 0.5506 \\
\textit{yuccatest}    & 30k & 18.53 & 0.4973 & 0.6467 \\
\textit{miningtunnel} & 30k & 11.85 & 0.3415 & 0.7327 \\
\textit{housecorner}  & 30k & 24.61 & 0.6255 & 0.5222 \\
\midrule
Mean                  & 30k & 19.21 & 0.5250 & 0.6155 \\
\bottomrule
\end{tabular}
\caption{Per-scene results for our vanilla Raw3DGS control: SH appearance with log-L1 and weighted-L2 supervision~\cite{mildenhall2022rawnerf}, trained for $30{,}000$ iterations. Metrics are computed in sRGB under the RawNeRF-style per-channel affine protocol; \textit{housecorner} is averaged over three reference views.}
\label{tab:raw3dgs-perscene}
\end{table}

\section{Full Matched-Input and Common-Output Results}
\label{app:matched-full}

\paragraph{Training configurations.}
RetinexGS and LE3D are trained for 15k and 30k iterations, respectively, and RawNeRF for 250k iterations. For LE3D and RawNeRF, we adapt the data loaders and camera intrinsics to the shared half-packed RAW inputs and retain the remaining hyperparameters from their released configurations.

\paragraph{Per-scene results.}
Table~\ref{tab:matched-full} reports complete per-scene values for the matched-input comparison in Sec.~\ref{sec:matched-input}. HDRSplat+PMRID retains its native PMRID pipeline and is included only as a common output-domain comparison.

\begin{table*}[h]
\centering
\scriptsize
\setlength{\tabcolsep}{3.5pt}
\resizebox{\textwidth}{!}{%
\begin{tabular}{llccccc}
\toprule
Scene & Method & sRGB PSNR & sRGB SSIM & sRGB LPIPS & RAW PSNR & RAW SSIM\\
\midrule
\multirow{4}{*}{\textit{pianotest}}
& \textbf{RetinexGS} & \textbf{25.84} & \textbf{0.623} & \textbf{0.461} & \textbf{62.74} & \textbf{0.9987}\\
& LE3D (half-packed) & 24.67 & 0.593 & 0.501 & 60.05 & 0.9979\\
& RawNeRF (half-packed) & 22.84 & 0.476 & 0.562 & 59.69 & 0.9978\\
& HDRSplat (native) & 22.84 & 0.565 & 0.490 & 59.19 & 0.9970\\
\midrule
\multirow{4}{*}{\textit{officetest}}
& \textbf{RetinexGS} & \textbf{25.58} & \textbf{0.635} & \textbf{0.440} & \textbf{61.30} & \textbf{0.9981}\\
& LE3D (half-packed) & 24.44 & 0.607 & 0.496 & 59.78 & 0.9973\\
& RawNeRF (half-packed) & 22.28 & 0.473 & 0.557 & 56.61 & 0.9954\\
& HDRSplat (native) & 23.97 & 0.587 & 0.475 & 57.64 & 0.9960\\
\midrule
\multirow{4}{*}{\textit{yuccatest}}
& \textbf{RetinexGS} & \textbf{23.91} & \textbf{0.560} & \textbf{0.509} & \textbf{61.12} & \textbf{0.9983}\\
& LE3D (half-packed) & 21.99 & 0.511 & 0.595 & 58.24 & 0.9965\\
& RawNeRF (half-packed) & 21.36 & 0.401 & 0.605 & 58.88 & 0.9973\\
& HDRSplat (native) & 19.73 & 0.396 & 0.638 & 55.92 & 0.9945\\
\midrule
\multirow{4}{*}{\textit{miningtunnel}}
& \textbf{RetinexGS} & \textbf{23.47} & \textbf{0.490} & \textbf{0.505} & \textbf{68.15} & \textbf{0.9995}\\
& LE3D (half-packed) & 20.59 & 0.414 & 0.613 & 66.56 & 0.9991\\
& RawNeRF (half-packed) & 22.06 & 0.438 & 0.536 & 67.28 & \textbf{0.9995}\\
& HDRSplat (native) & 15.78 & 0.334 & 0.733 & 61.43 & 0.9968\\
\midrule
\multirow{4}{*}{\textit{housecorner}}
& \textbf{RetinexGS} & \textbf{26.27} & \textbf{0.692} & \textbf{0.440} & \textbf{63.63} & \textbf{0.9990}\\
& LE3D (half-packed) & 22.48 & 0.623 & 0.560 & 61.07 & 0.9975\\
& RawNeRF (half-packed) & 23.00 & 0.633 & 0.491 & 62.12 & 0.9983\\
& HDRSplat (native) & 16.57 & 0.517 & 0.755 & 52.38 & 0.9840\\
\midrule
\multirow{4}{*}{Mean}
& \textbf{RetinexGS} & \textbf{25.01} & \textbf{0.600} & \textbf{0.471} & \textbf{63.39} & \textbf{0.9987}\\
& LE3D (half-packed) & 22.83 & 0.550 & 0.553 & 61.14 & 0.9977\\
& RawNeRF (half-packed) & 22.31 & 0.484 & 0.550 & 60.92 & 0.9977\\
& HDRSplat (native) & 19.78 & 0.480 & 0.618 & 57.31 & 0.9937\\
\bottomrule
\end{tabular}
}
\caption{Complete five-scene values used in the matched-input and common-output analyses. RetinexGS, LE3D, and RawNeRF receive identical half-packed tensors; HDRSplat retains its native PMRID pipeline and is included only as a common output-domain comparison. sRGB reports PSNR/SSIM/LPIPS under the RawNeRF-style affine protocol; RAW reports PSNR/SSIM without affine fitting.}
\label{tab:matched-full}
\end{table*}

\paragraph{Native-pipeline RAW diagnostics.}
Table~\ref{tab:raw-diagnostics} retains the native-domain RAW values from the public comparison for completeness. Because these methods differ in Bayer sampling, demosaicing, scaling, and supervision, we do not treat the values as a controlled ranking; the matched-input RAW metrics above serve that purpose.

\begin{table}[h]
\centering
\small
\begin{tabular}{lcc}
\toprule
Method & RAW PSNR$\uparrow$ & RAW SSIM$\uparrow$ \\
\midrule
HDR-3DGS & 56.50 & 0.9926 \\
RawNeRF & 58.69 & 0.9969 \\
RawGS & 59.28 & 0.9971 \\
LE3D & 61.08 & 0.9983 \\
RetinexGS & 61.72 & 0.9984 \\
\bottomrule
\end{tabular}
\caption{Method-specific RAW-domain diagnostics on the RawNeRF test scenes. These values are documented but not used as the primary cross-method ranking.}
\label{tab:raw-diagnostics}
\end{table}

\section{Full Composition Results}
\label{app:placement-full}

Table~\ref{tab:placement-full} reports per-scene results for the composition comparison in Sec.~\ref{sec:composition-efficiency}.

\begin{table*}[h]
\centering
\scriptsize
\setlength{\tabcolsep}{3.8pt}
\begin{tabular}{llccccc}
\toprule
Scene & Composition & sRGB PSNR$\uparrow$ & SSIM$\uparrow$ & LPIPS$\downarrow$
      & RAW PSNR$\uparrow$ & RAW SSIM$\uparrow$ \\
\midrule
\multirow{2}{*}{\textit{pianotest}}
 & Primitive-level & 25.84 & 0.6230 & 0.4610 & 62.74 & 0.9987 \\
 & Image-space     & 25.72 & 0.6215 & 0.4635 & 62.59 & 0.9986 \\
\multirow{2}{*}{\textit{officetest}}
 & Primitive-level & 25.58 & 0.6350 & 0.4400 & 61.30 & 0.9981 \\
 & Image-space     & 25.48 & 0.6345 & 0.4437 & 61.02 & 0.9980 \\
\multirow{2}{*}{\textit{yuccatest}}
 & Primitive-level & 23.91 & 0.5600 & 0.5090 & 61.12 & 0.9983 \\
 & Image-space     & 23.77 & 0.5596 & 0.5149 & 60.94 & 0.9983 \\
\multirow{2}{*}{\textit{miningtunnel}}
 & Primitive-level & 23.47 & 0.4897 & 0.5053 & 68.15 & 0.9995 \\
 & Image-space     & 23.69 & 0.4896 & 0.5048 & 67.77 & 0.9995 \\
\multirow{2}{*}{\textit{housecorner}}
 & Primitive-level & 26.27 & 0.6919 & 0.4401 & 63.63 & 0.9990 \\
 & Image-space     & 26.13 & 0.6906 & 0.4446 & 63.67 & 0.9990 \\
\midrule
\multirow{2}{*}{Five-scene mean}
 & Primitive-level & 25.01 & 0.6000 & 0.4710 & 63.39 & 0.9987 \\
 & Image-space     & 24.96 & 0.5992 & 0.4743 & 63.20 & 0.9987 \\
\bottomrule
\end{tabular}
\caption{Full composition comparison. Each model is rendered with its requested operator, and the two arms otherwise use the same factor parameterization, data, optimization, and evaluation. Primitive-level entries are the canonical RetinexGS checkpoints used throughout the main results and ablations; image-space entries are the independently verified comparison runs. Primitive-level composition requires approximately half the wall-clock training time (Sec.~\ref{sec:composition-efficiency}).}
\label{tab:placement-full}
\end{table*}

\section{Runtime and Rendering Efficiency}
\label{app:efficiency}

\paragraph{Protocol.}
We measure rendering at $2016\times1512$ with fixed geometry, opacity, camera views, Gaussian counts, and distortion-aware rasterization. The SH attributes in the matched cost control are evaluation-only, so Table~\ref{tab:efficiency} measures throughput and peak memory rather than reconstruction quality. Composition training time is measured over the complete 15k-iteration optimization under the same hardware, logging, and evaluation settings, and is reported relative to the image-space arm in Sec.~\ref{sec:composition-efficiency}.

\begin{table}[h]
    \centering
    \small
    \setlength{\tabcolsep}{5pt}
    \begin{tabular}{lccccc}
        \toprule
        \multirow{2}{*}{Scene}
        & \multirow{2}{*}{\# Gaussians}
        & \multicolumn{2}{c}{RetinexGS}
        & \multicolumn{2}{c}{Matched SH control} \\
        \cmidrule(lr){3-4}
        \cmidrule(lr){5-6}
        & & FPS$\uparrow$ & Memory (MB)$\downarrow$
        & FPS$\uparrow$ & Memory (MB)$\downarrow$ \\
        \midrule
        \textit{pianotest}
        & 340k & 91 & 602 & 148 & 614--661 \\
        \textit{housecorner}
        & 138k & 145 & 471 & 176--187 & 475--494 \\
        \textit{miningtunnel}
        & 225k & 138 & 427 & 207--215 & 434--464 \\
        \bottomrule
    \end{tabular}
    \caption{\textbf{Rendering efficiency.} Throughput and peak memory at $2016\times1512$ resolution on an NVIDIA RTX 4090. SH ranges span degrees 1 and 3. Geometry, opacity, cameras, Gaussian counts, and the distortion-aware rasterizer are fixed to isolate appearance-evaluation cost.}
    \label{tab:efficiency}
\end{table}

\section{Additional Qualitative Results}
\label{app:ourscenes}

\begin{figure*}[t]
    \centering
    \includegraphics[width=\textwidth]{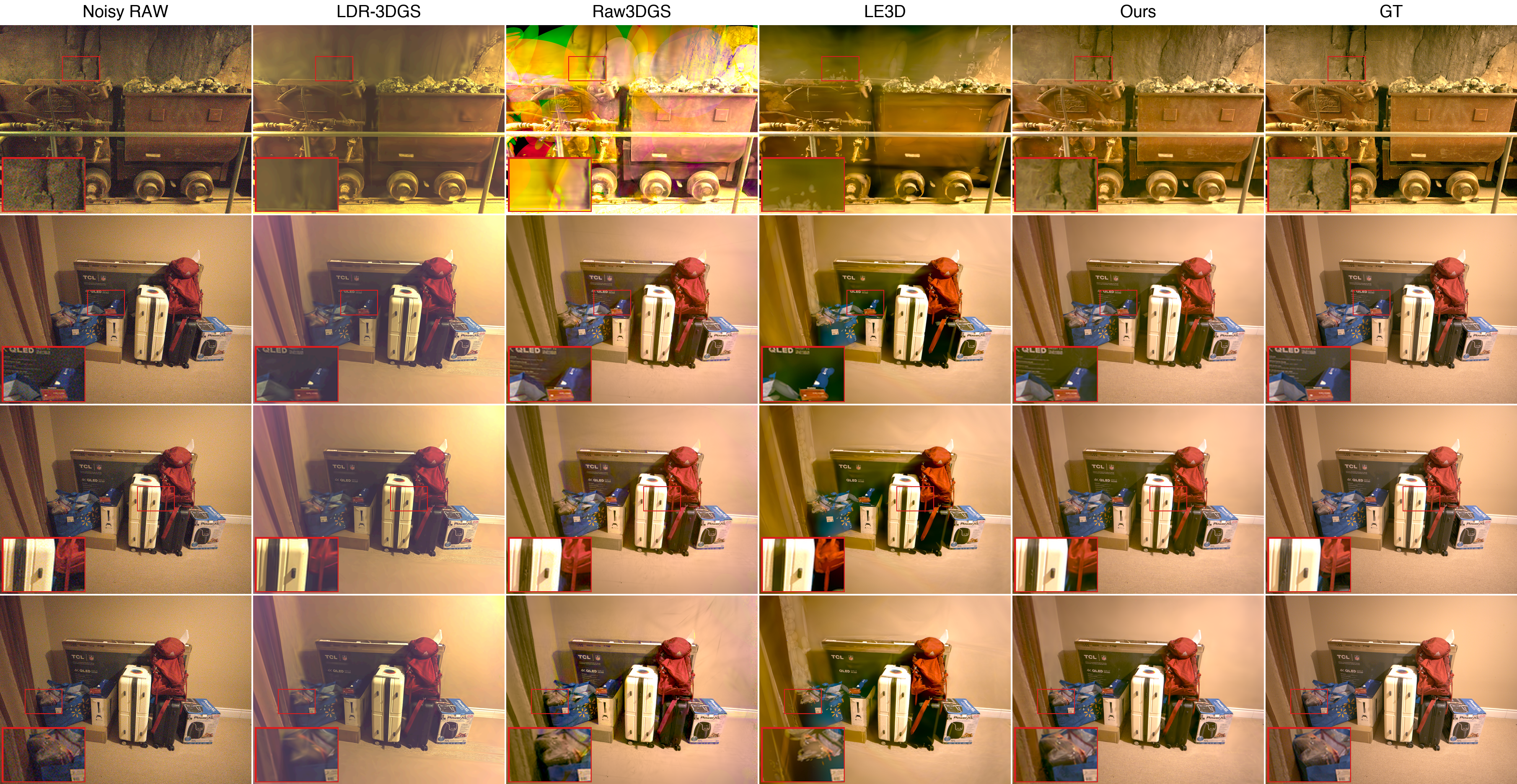}
    \caption{\textbf{Qualitative comparison on our self-collected benchmark.} Test-view renderings on \textit{miningtunnel} and \textit{housecorner}, where long-exposure tripod-stabilized reference views are available. From left to right: noisy RAW input (same pose as GT, shown for reference), LDR-3DGS~\cite{kerbl20233dgs}, Raw3DGS (our impl.), LE3D~\cite{jin2024le3d}, RetinexGS (ours), and GT reference. All RAW-based renderings are aligned to GT via per-channel affine correction in linear RAW; LDR-3DGS is aligned in sRGB. Red boxes indicate zoomed regions. Our method produces fewer floating artifacts, more accurate color, and better details in dark regions.}
    \label{fig:our-comparison}
\end{figure*}

Fig.~\ref{fig:our-comparison} provides per-method renderings on \textit{miningtunnel} and \textit{housecorner}. Compared with the RawNeRF scenes, the gap to the baselines is larger on these captures, particularly in textureless regions and on fine details such as label text.

\section{DNG-Informed Weighting vs.\ Weighted L2}
\label{app:nll_vs_l2}

\begin{figure}[h]
    \centering
    \includegraphics[width=\linewidth]{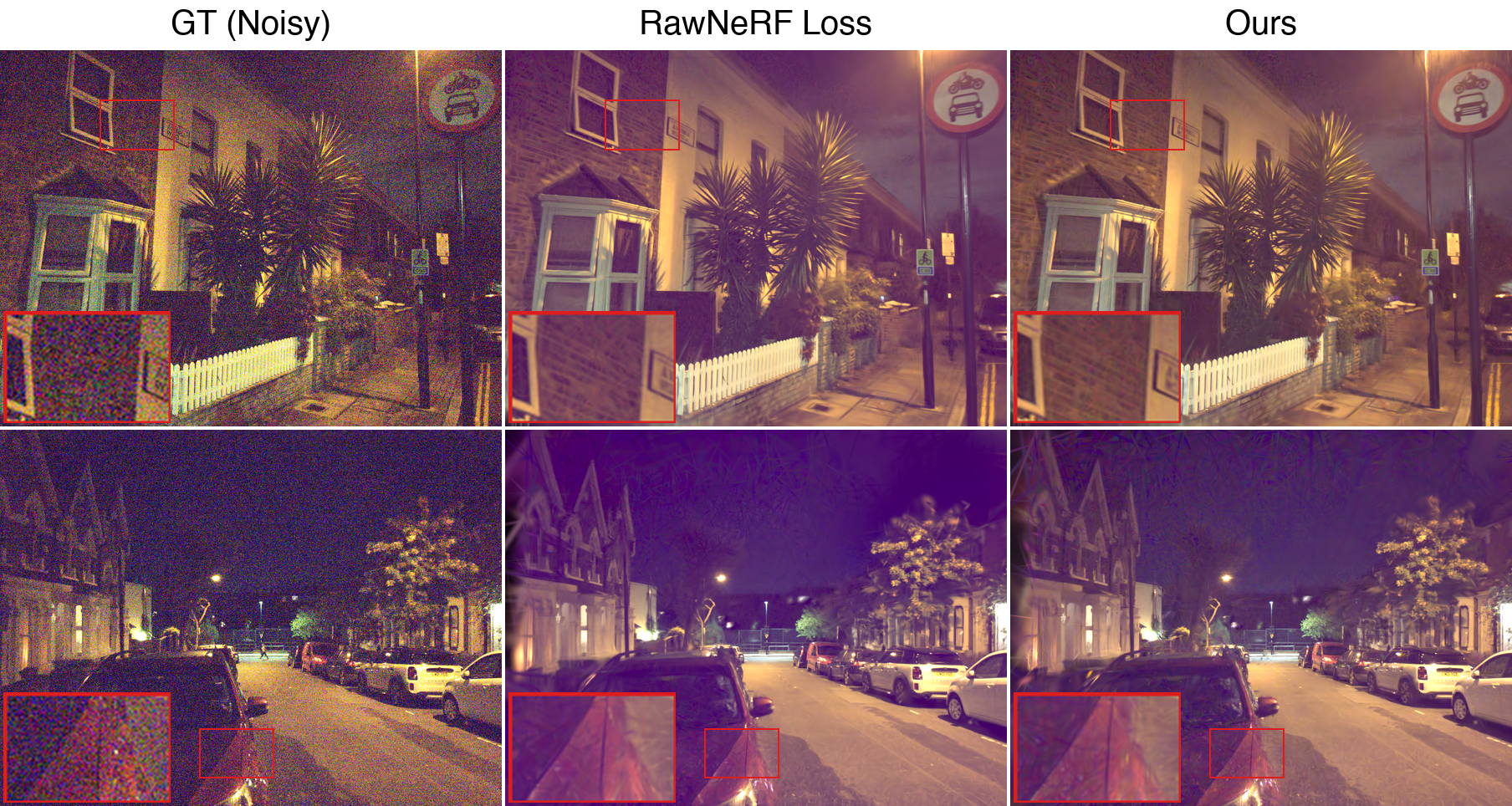}
    \caption{Replacing our DNG-informed weighting with RawNeRF's weighted L2 loss introduces a visible purple cast in low-signal regions, despite nearly identical PSNR/SSIM in Table~\ref{tab:ablation}.}
    \label{fig:color_shift}
\end{figure}

Fig.~\ref{fig:color_shift} shows a purple cast in a low-signal region when the DNG-informed term is replaced by RawNeRF's weighted L2 loss~\cite{mildenhall2022rawnerf}; LPIPS changes more clearly than PSNR/SSIM in this example. Our weighting accounts for channel-specific $(a_c,b_c)$ and the variance reduction from averaging two green photosites. The figure is qualitative evidence rather than a calibrated color-accuracy study. A mild residual shift remains in our result, suggesting that DNG metadata may not perfectly model capture-specific noise.

\end{document}